%% file: style.tex
\documentclass{article}
\usepackage{ijcai17}

\usepackage{times}

\usepackage{times}
\usepackage{epsfig}
\usepackage{graphicx}
\usepackage{amsmath}
\usepackage{amssymb}
\usepackage{subfigure}
\usepackage{scrextend}
\usepackage{enumerate}
\usepackage{overpic}
\usepackage[table]{xcolor}

\newcommand{\col}[2] {#1_{\cdot #2}}

\usepackage[small]{caption}

\usepackage[breaklinks=true,bookmarks=false]{hyperref}

\begin{document}

\title{Demystifying Neural Style Transfer}

\author{Yanghao Li$^\dagger$~~~~~ Naiyan Wang$^\ddagger$~~~~~ Jiaying Liu$^\dagger$\thanks{Corresponding author}~~~~~ Xiaodi Hou$^\ddagger$\\
$^\dagger$ Institute of Computer Science and Technology, Peking University~~~~~\\
$^\ddagger$ TuSimple\\
{ lyttonhao@pku.edu.cn}~~{ winsty@gmail.com}~~
{ liujiaying@pku.edu.cn}~~
{ xiaodi.hou@gmail.com}
}

\maketitle
\rowcolors{2}{white}{gray!25}

\graphicspath{{figures/}}

\input{src/abstract.tex}

\input{src/introduction.tex}

\input{src/related.tex}

\input{src/proposed.tex}

\input{src/experiments.tex}

\input{src/conclusion.tex}

\section*{Acknowledgement} This work was supported by the National Natural Science Foundation of China under Contract 61472011.

{
\bibliographystyle{named}
\bibliography{egbib}
}

\end{document}

%% file: src/abstract.tex
\begin{abstract}

Neural Style Transfer~\cite{neuralart} has recently demonstrated very exciting results which catches eyes in both academia and industry. Despite the amazing results, the principle of neural style transfer, especially why the Gram matrices could represent style remains unclear. In this paper, we propose a novel interpretation of neural style transfer by treating it as a domain adaptation problem. Specifically, we theoretically show that matching the Gram matrices of feature maps is equivalent to minimize the Maximum Mean Discrepancy (MMD) with the second order polynomial kernel. Thus, we argue that the essence of neural style transfer is to match the feature distributions between the style images and the generated images. To further support our standpoint, we experiment with several other distribution alignment methods, and achieve appealing results. We believe this novel interpretation connects these two important research fields, and could enlighten future researches.

\end{abstract}

%% file: src/introduction.tex
\begin{section}{Introduction}

Transferring the style from one image to another image is an interesting yet difficult problem. There have been many efforts to develop efficient methods for automatic style transfer~\cite{hertzmann2001image,efros2001image,efros1999texture,shih2014style,kwatra2005texture}. Recently, Gatys \emph{et al.} proposed a seminal work~\cite{neuralart}: It captures the style of artistic images and transfer it to other images using Convolutional Neural Networks (CNN). This work formulated the problem as finding an image that matching both the content and style statistics based on the neural activations of each layer in CNN. It achieved impressive results and several follow-up works improved upon this innovative approaches~\cite{johnson2016perceptual,ulyanov2016texture,ruder2016artistic,ledig2016photo}. Despite the fact that this work has drawn lots of attention, the fundamental element of style representation: the Gram matrix in~\cite{neuralart} is not fully explained. The reason why Gram matrix can represent artistic style still remains a mystery.

In this paper, we propose a novel interpretation of neural style transfer by casting it as a special domain adaptation~\cite{beijbom2012domain,patel2015visual} problem. We theoretically prove that matching the Gram matrices of the neural activations can be seen as minimizing a specific Maximum Mean Discrepancy (MMD)~\cite{mmd}. This reveals that neural style transfer is intrinsically a process of distribution alignment of the neural activations between images. Based on this illuminating analysis, we also experiment with other distribution alignment methods, including MMD with different kernels and a simplified moment matching method. These methods achieve diverse but all reasonable style transfer results. Specifically, a transfer method by MMD with linear kernel achieves comparable visual results yet with a lower complexity. Thus, the second order interaction in Gram matrix is not a must for style transfer. Our interpretation provides a promising direction to design style transfer methods with different visual results. To summarize, our contributions are shown as follows:
\begin{enumerate}
\item First, we demonstrate that matching Gram matrices in neural style transfer~\cite{neuralart} can be reformulated as minimizing  MMD with the second order polynomial kernel.
\item Second, we extend the original neural style transfer with different distribution alignment methods based on our novel interpretation.
\end{enumerate}

\end{section}

%% file: src/related.tex
\begin{section}{Related Work}
In this section, we briefly review some closely related works and the key concept MMD in our interpretation.
\begin{paragraph}{Style Transfer}
Style transfer is an active topic in both academia and industry. Traditional methods mainly focus on the non-parametric patch-based texture synthesis and transfer, which resamples pixels or patches from the original source texture images~\cite{hertzmann2001image,efros2001image,efros1999texture,liang2001real}. Different methods were proposed to improve the quality of the patch-based synthesis and constrain the structure of the target image. For example, the image quilting algorithm based on dynamic programming was proposed to find optimal texture boundaries in~\cite{efros2001image}. A Markov Random Field (MRF) was exploited to preserve global texture structures in~\cite{frigo2016split}. However, these non-parametric methods suffer from a fundamental limitation that they only use the low-level features of the images for transfer. 

Recently, neural style transfer~\cite{neuralart} has demonstrated remarkable results for image stylization. It fully takes the advantage of the powerful representation of Deep Convolutional Neural Networks (CNN). This method used Gram matrices of the neural activations from different layers of a CNN to represent the artistic style of a image. Then it used an iterative optimization method to generate a new image from white noise by matching the neural activations with the content image and the Gram matrices with the style image. This novel technique attracts many follow-up works for different aspects of improvements and applications. To speed up the iterative optimization process in~\cite{neuralart}, Johnson \emph{et al.}~\cite{johnson2016perceptual} and Ulyanov \emph{et al.}~\cite{ulyanov2016texture} trained a feed-forward generative network for fast neural style transfer. 
\textcolor{black}{To improve the transfer results in~\cite{neuralart}, different complementary schemes are proposed, including spatial constraints~\cite{selim2016painting}, semantic guidance~\cite{neuraldoodle} and Markov Random Field (MRF) prior~\cite{li2016combining}. There are also some extension works to apply neural style transfer to other applications. Ruder \emph{et al.}~\cite{ruder2016artistic} incorporated temporal consistence terms by penalizing deviations between frames for video style transfer. Selim \emph{et al.}~\cite{selim2016painting} proposed novel spatial constraints through gain map for portrait painting transfer. }
Although these methods further improve over the original neural style transfer, they all ignore the fundamental question in neural style transfer: \emph{Why could the Gram matrices represent the artistic style?} This vagueness of the understanding limits the further research on the neural style transfer. 
\end{paragraph}

\begin{paragraph}{Domain Adaptation}
Domain adaptation belongs to the area of transfer learning~\cite{pan2010survey}. It aims to transfer the model that is learned on the source domain to the unlabeled target domain. The key component of domain adaptation is to measure and minimize the difference between source and target distributions. The most common discrepancy metric is Maximum Mean Discrepancy (MMD)~\cite{mmd}, which measure the difference of sample mean in a Reproducing Kernel Hilbert Space. It is a popular choice in domain adaptation works~\cite{ddc,dan,long2016unsupervised}. Besides MMD, Sun \emph{et al.}~\cite{coral} aligned the second order statistics by whitening the data in source domain and then re-correlating to the target domain. In \cite{adabn}, Li \emph{et al.} proposed a parameter-free deep adaptation method by simply modulating the statistics in all Batch Normalization (BN) layers.
\end{paragraph}

\begin{paragraph}{Maximum Mean Discrepancy} Suppose there are two sets of samples $X=\{\mathbf{x}_i\}_{i=1}^{n}$ and $Y = \{\mathbf{y}_j\}_{j=1}^{m}$ where $\mathbf{x}_i$ and $\mathbf{y}_j$ are generated from distributions $p$ and $q$, respectively. Maximum Mean Discrepancy (MMD) is a popular test statistic for the two-sample testing problem, where acceptance or rejection decisions are made for a null hypothesis $p = q$~\cite{mmd}. Since the population MMD vanishes if and only $p = q$, the MMD statistic can be used to measure the difference between two distributions. Specifically, we calculates MMD defined by the difference between the mean embedding on the two sets of samples. Formally, the squared MMD is defined as:
\begin{small}
\begin{equation}\label{mmd}
\begin{aligned}
&  \text{MMD}^2[X, Y]\\
		 = ~ &\| \mathbf{E}_x[\phi(\mathbf{x})] - \mathbf{E}_y[\phi(\mathbf{y})] \|^2\\
		= ~&\| \frac{1}{n}\sum_{i=1}^{n}\phi(\mathbf{x}_i) - \frac{1}{m}\sum_{j=1}^{m}\phi(\mathbf{y}_j) \|^2\\
		= ~&\frac{1}{n^2}\sum_{i=1}^{n}\sum_{i'=1}^{n}\phi(\mathbf{x}_i)^T\phi(\mathbf{x}_{i'}) + 
		   \frac{1}{m^2}\sum_{j=1}^{m}\sum_{j'=1}^{m}\phi(\mathbf{y}_j)^T\phi(\mathbf{y}_{j'}) \\
		&   -\frac{2}{nm}\sum_{i=1}^{n}\sum_{j=1}^{m}\phi(\mathbf{x}_i)^T\phi(\mathbf{y}_{j}),
\end{aligned}
\end{equation}
\end{small}
where $\phi(\cdot)$ is the explicit feature mapping function of MMD. Applying the associated kernel function $k(\mathbf{x}, \mathbf{y}) = \langle\phi(\mathbf{x}), \phi(\mathbf{y})\rangle$, the Eq.~\ref{mmd} can be expressed in the form of kernel:
\begin{small}
\begin{equation}{\label{mmd_kernel}}
\begin{aligned}
&\text{MMD}^2[X, Y]\\
	= ~ & \frac{1}{n^2}\sum_{i=1}^{n}\sum_{i'=1}^{n}k(\mathbf{x}_i, \mathbf{x}_{i'}) + 
		   \frac{1}{m^2}\sum_{j=1}^{m}\sum_{j'=1}^{m}k(\mathbf{y}_j, \mathbf{y}_{j'}) \\
	&	   -\frac{2}{nm}\sum_{i=1}^{n}\sum_{j=1}^{m}k(\mathbf{x}_i, \mathbf{y}_j).
\end{aligned}
\end{equation}
\end{small}
The kernel function $k(\cdot, \cdot)$ implicitly defines a mapping to a higher dimensional feature space.
\end{paragraph}

\end{section}

%% file: src/proposed.tex
\begin{section}{Understanding Neural Style Transfer}
In this section, we first theoretically demonstrate that matching Gram matrices is equivalent to minimizing a specific form of MMD. Then based on this interpretation, we extend the original neural style transfer with different distribution alignment methods.

Before explaining our observation, we first briefly review the original neural style transfer approach~\cite{neuralart}. The goal of style transfer is to generate a stylized image $\mathbf{x}^*$ given a content image $\mathbf{x}_c$ and a reference style image $\mathbf{x}_s$. The feature maps of $\mathbf{x}^*$, $\mathbf{x}_c$ and $\mathbf{x}_s$ in the layer $l$ of a CNN are denoted by $\mathbf{F}^l \in \mathbb{R}^{N_l \times M_l}$, $\mathbf{P}^l \in \mathbb{R}^{N_l \times M_l}$ and $\mathbf{S}^l \in \mathbb{R}^{N_l \times M_l}$ respectively, where $N_l$ is the number of the feature maps in the layer $l$ and $M_l$ is the height times the width of the feature map.

In \cite{neuralart}, neural style transfer iteratively generates $\mathbf{x}^*$ by optimizing a content loss and a style loss:
\begin{equation}\label{eq:total_loss}
\begin{aligned}
\mathcal{L} = \alpha\mathcal{L}_{content} + \beta\mathcal{L}_{style},
\end{aligned}
\end{equation}
where $\alpha$ and $\beta$ are the weights for content and style losses, $\mathcal{L}_{content}$ is defined by the squared error between the feature maps of a specific layer $l$ for $\mathbf{x}^*$ and $\mathbf{x}_c$:
\begin{equation}
\begin{aligned}
\mathcal{L}_{content} = \frac{1}{2}\sum_{i=1}^{N_l}\sum_{j=1}^{M_l}(F_{ij}^l - P_{ij}^l)^2,
\end{aligned}
\end{equation}
and $\mathcal{L}_{style}$ is the sum of several style loss $\mathcal{L}_{style}^{l}$ in different layers:
\begin{equation}
\begin{aligned}
\mathcal{L}_{style} = \sum_{l} w_l\mathcal{L}_{style}^{l},
\end{aligned}
\end{equation}
where $w_l$ is the weight of the loss in the layer $l$ and  $\mathcal{L}_{style}^{l}$ is defined by the squared error between the features correlations expressed by Gram matrices of $\mathbf{x}^*$ and $\mathbf{x}_s$:
\begin{equation}\label{eq_style}
\begin{aligned}
\mathcal{L}_{style}^l = \frac{1}{4N_l^2M_l^2}\sum_{i=1}^{N_l}\sum_{j=1}^{N_l}(G_{ij}^l - A_{ij}^l)^2,
\end{aligned}
\end{equation}
where the Gram matrix $\mathbf{G}^l \in \mathbb{R}^{N_l \times N_l}$ is the inner product between the vectorized feature maps of $\mathbf{x}^*$ in layer $l$:
\begin{equation}
\begin{aligned}
G_{ij}^l = \sum_{k=1}^{M_l}F_{ik}^lF_{jk}^l,
\end{aligned}
\end{equation}
and similarly $\mathbf{A}^l$ is the Gram matrix corresponding to $\mathbf{S}^l$.

\begin{figure*}[hbtp]
\begin{footnotesize}
\begin{equation}\label{eq:provemmd}
\begin{aligned}
&\mathcal{L}_{style}^l = \frac{1}{4N_l^2M_l^2}\sum_{i=1}^{N_l}\sum_{j=1}^{N_l}(\sum_{k=1}^{M_l}F_{ik}^lF_{jk}^l - \sum_{k=1}^{M_l}S_{ik}^lS_{jk}^l)^2\\
&=  \frac{1}{4N_l^2M_l^2}\sum_{i=1}^{N_l}\sum_{j=1}^{N_l}\Big( 
		(\sum_{k=1}^{M_l}F_{ik}^lF_{jk}^l)^2 +
		(\sum_{k=1}^{M_l}S_{ik}^lS_{jk}^l)^2 -
		2(\sum_{k=1}^{M_l}F_{ik}^lF_{jk}^l)(\sum_{k=1}^{M_l}S_{ik}^lS_{jk}^l) \Big)\\
&=  \frac{1}{4N_l^2M_l^2}\sum_{i=1}^{N_l}\sum_{j=1}^{N_l}\sum_{k_1=1}^{M_l}\sum_{k_2=1}^{M_l} ( F_{ik_1}^l F_{jk_1}^l F_{ik_2}^l F_{jk_2}^l + 
S_{ik_1}^l S_{jk_1}^l S_{ik_2}^l S_{jk_2}^l - 2 F_{ik_1}^l F_{jk_1}^l  S_{ik_2}^l S_{jk_2}^l)\\
&= \frac{1}{4N_l^2M_l^2}\sum_{k_1=1}^{M_l}\sum_{k_2=1}^{M_l} \sum_{i=1}^{N_l}\sum_{j=1}^{N_l} ( F_{ik_1}^l F_{jk_1}^l F_{ik_2}^l F_{jk_2}^l + 
S_{ik_1}^l S_{jk_1}^l S_{ik_2}^l S_{jk_2}^l - 2 F_{ik_1}^l F_{jk_1}^l  S_{ik_2}^l S_{jk_2}^l)\\
&= \frac{1}{4N_l^2M_l^2}\sum_{k_1=1}^{M_l}\sum_{k_2=1}^{M_l} 
   \Big( (\sum_{i=1}^{N_l}F_{ik_1}^lF_{ik_2}^l)^2 + 
         (\sum_{i=1}^{N_l}S_{ik_1}^lS_{ik_2}^l)^2 - 
         2(\sum_{i=1}^{N_l}F_{ik_1}^lS_{ik_2}^l)^2
   \Big)\\
&= \frac{1}{4N_l^2M_l^2}\sum_{k_1=1}^{M_l} \sum_{k_2=1}^{M_l} 
   \Big(  ({\col{\mathbf{f}^l}{k_1}}^T \col{\mathbf{f}^l}{k_2} )^2  + 
   		  ({\col{\mathbf{s}^l}{k_1}}^T \col{\mathbf{s}^l}{k_2} )^2  -
   		  2 ({\col{\mathbf{f}^l}{k_1}}^T \col{\mathbf{s}^l}{k_2} )^2 
   \Big),
\end{aligned}
\end{equation}
\end{footnotesize}
\vspace{-2mm}
\end{figure*}

\begin{subsection}{Reformulation of the Style Loss}
In this section, we reformulated the style loss $\mathcal{L}_{style}$ in Eq.~\ref{eq_style}. By expanding the Gram matrix in Eq.~\ref{eq_style}, we can get the formulation of Eq.~\ref{eq:provemmd}, where $\col{\mathbf{f}^l}{k}$ and $\col{\mathbf{s}^l}{k}$ is the $k$-th column of $\mathbf{F}^l$ and $\mathbf{S}^l$.

By using the second order degree polynomial kernel $k(\mathbf{x}, \mathbf{y}) = (\mathbf{x}^T\mathbf{y})^2$, Eq.~\ref{eq:provemmd} can be represented as:
\begin{equation}\label{eq:prove_result}
\begin{aligned}
\mathcal{L}_{style}^l =& \frac{1}{4N_l^2M_l^2}\sum_{k_1=1}^{M_l}\sum_{k_2=1}^{M_l}  
	\Big( k(\col{\mathbf{f}^l}{k_1}, \col{\mathbf{f}^l}{k_2}) \\
		 & + k(\col{\mathbf{s}^l}{k_1}, \col{\mathbf{s}^l}{k_2}) 
		 - 2k(\col{\mathbf{f}^l}{k_1}, \col{\mathbf{s}^l}{k_2})
	\Big)\\
	=& \frac{1}{4N_l^2} \text{MMD}^2[\mathcal{F}^{l}, \mathcal{S}^{l}],
\end{aligned}
\end{equation}
where $\mathcal{F}^{l}$ is the feature set of $\mathbf{x}^*$ where each sample is a column of $\mathbf{F}^l$, and $\mathcal{S}^{l}$ corresponds to the style image $\mathbf{x}_s$. In this way, the activations at each position of feature maps is considered as an individual sample. Consequently, the style loss ignores the positions of the features, which is desired for style transfer. In conclusion, the above reformulations suggest two important findings:
\begin{enumerate}
\item The style of a image can be intrinsically represented by feature distributions in different layers of a CNN.
\item The style transfer can be seen as a distribution alignment process from the content image to the style image. 
\end{enumerate}
\end{subsection}

\begin{subsection}{Different Adaptation Methods for Neural Style Transfer}\label{sec:methods}
Our interpretation reveals that neural style transfer can be seen as a problem of distribution alignment, which is also at the core in domain adaptation. If we consider the style of one image in a certain layer of CNN as a ``domain'', style transfer can also be seen as a special domain adaptation problem. The specialty of this problem lies in that we treat the feature at each position of feature map as one individual data sample, instead of that in traditional domain adaptation problem in which we treat each image as one data sample. (\emph{e.g.} The feature map of the last convolutional layer in VGG-19 model is of size $14 \times 14$, then we have totally 196 samples in this ``domain''.)


Inspired by the studies of domain adaptation, we extend neural style transfer with different adaptation methods in this subsection.

\begin{paragraph}{MMD with Different Kernel Functions}
As shown in Eq.~\ref{eq:prove_result}, matching Gram matrices in neural style transfer can been seen as a MMD process with second order polynomial kernel. It is very natural to apply other kernel functions for MMD in style transfer. First, if using MMD statistics to measure the style discrepancy, the style loss can be defined as:
\begin{equation}\label{eq:style_mmd}
\begin{aligned}
&\mathcal{L}_{style}^l = \frac{1}{Z^l_k}\text{MMD}^2[\mathcal{F}^{l}, \mathcal{S}^{l}],\\
	&= \frac{1}{Z^l_k}\sum_{i=1}^{M_l}\sum_{j=1}^{M_l}\Big( 
		k(\col{\mathbf{f}^l}{i}, \col{\mathbf{f}^l}{j}) +
		  k(\col{\mathbf{s}^l}{i}, \col{\mathbf{s}^l}{j}) - 2k(\col{\mathbf{f}^l}{i}, \col{\mathbf{s}^l}{j})
	\Big),
\end{aligned}
\end{equation}
where $Z^l_k$ is the normalization term corresponding to different scale of the feature map in the layer $l$ and the choice of  kernel function. Theoretically, different kernel function implicitly maps features to different higher dimensional space. Thus, we believe that different kernel functions should capture different aspects of a style. We adopt the following three popular kernel functions in our experiments:
\begin{enumerate}[{(1)}]
	\item Linear kernel: $k(\mathbf{x}, \mathbf{y}) = \mathbf{x}^T\mathbf{y}$;
	\item Polynomial kernel: $k(\mathbf{x}, \mathbf{y}) = (\mathbf{x}^T\mathbf{y} + c)^d$;
	\item Gaussian kernel: $k(\mathbf{x}, \mathbf{y}) = \exp\big( -\frac{\|\mathbf{x} - \mathbf{y}\|_2^2}{2\sigma^2}  \big)$.
\end{enumerate}
For polynomial kernel, we only use the version with $d = 2$. Note that matching Gram matrices is equivalent to the polynomial kernel with $c = 0$ and $d = 2$. For the Gaussian kernel, we adopt the unbiased estimation of MMD~\cite{gretton2012optimal}, which samples $M_l$ pairs in Eq.~\ref{eq:style_mmd} and thus can be computed with linear complexity. 

\end{paragraph}

\begin{paragraph}{BN Statistics Matching} In~\cite{adabn}, the authors found that the statistics (\emph{i.e.} mean and variance) of Batch Normalization (BN) layers contains the traits of different domains. Inspired by this observation, they utilized separate BN statistics for different domain. This simple operation aligns the different domain distributions effectively. As a special domain adaptation problem, we believe that BN statistics of a certain layer can also represent the style. Thus, we construct another style loss by aligning the BN statistics (mean and standard deviation) of two feature maps between two images:
\begin{equation}
\begin{aligned}
&\mathcal{L}_{style}^l = \frac{1}{N_l} \sum_{i=1}^{N_l}\Big( ( \mu^i_{F^l} - \mu^i_{S^l} )^2 + ( \sigma^i_{F^l} - \sigma^i_{S^l} )^2 \Big),
\end{aligned}
\end{equation}
where $\mu^i_{F^l}$ and $\sigma^i_{F^l}$ is the mean and standard deviation of the $i$-th feature channel among all the positions of the feature map in the layer $l$ for image $\mathbf{x}^*$:
\begin{equation}
\begin{aligned}
\mu^i_{F^l} = \frac{1}{M_l} \sum_{j=1}^{M_l} F^l_{ij}, \quad
{\sigma^i_{F^l}}^2 =  \frac{1}{M_l} \sum_{j=1}^{M_l} (F^l_{ij} - \mu^i_{F^l})^2,
\end{aligned}
\end{equation}
and $\mu^i_{S^l}$ and $\sigma^i_{S^l}$ correspond to the style image $\mathbf{x}_s$.
\end{paragraph}

The aforementioned style loss functions are all differentiable and thus the style matching problem can be solved by back propagation iteratively.
\end{subsection}

\end{section}

%% file: src/experiments.tex
\begin{section}{Results}
In this section, we briefly introduce some implementation details and present results by our extended neural style transfer methods. Furthermore, we also show the results of fusing different neural style transfer methods, which combine different style losses. In the following, we refer the four extended style transfer methods introduced in Sec.~\ref{sec:methods} as \emph{linear}, \emph{poly}, \emph{Gaussian} and \emph{BN}, respectively. The images in the experiments are collected from the public implementations of neural style transfer\footnote{{\label{fn:mxnet-neural}{https://github.com/dmlc/mxnet/tree/master/example/neural-style}}}\footnote{{{https://github.com/jcjohnson/neural-style}}}\footnote{{{https://github.com/jcjohnson/fast-neural-style}}}.

\begin{paragraph}{Implementation Details}
In the implementation, we use the VGG-19 network~\cite{vgg} following the choice in~\cite{neuralart}. We also adopt the \emph{relu4\_2} layer for the content loss, and \emph{relu1\_1}, \emph{relu2\_1}, \emph{relu3\_1}, \emph{relu4\_1}, \emph{relu5\_1} for the style loss. The default weight factor $w_l$ is set as 1.0 if it is not specified. The target image $\mathbf{x}^*$ is initialized randomly and optimized iteratively until the relative change between successive iterations is under 0.5\%. The maximum number of iterations is set as 1000. For the method with Gaussian kernel MMD, the kernel bandwidth $\sigma^2$ is fixed as the mean of squared $l_2$ distances of the sampled pairs since it does not affect a lot on the visual results. Our implementation is based on the MXNet~\cite{mxnet} implementation\footref{fn:mxnet-neural} which reproduces the results of original neural style transfer~\cite{neuralart}.

Since the scales of the gradients of the style loss differ for different methods, and the weights $\alpha$ and $\beta$ in Eq.~\ref{eq:total_loss} affect the results of style transfer, we fix some factors to make a fair comparison. Specifically, we set $\alpha=1$ because the content losses are the same among different methods. Then, for each method, we first manually select a proper $\beta'$ such that the gradients on the $\mathbf{x}^*$ from the style loss are of the same order of magnitudes as those from the content loss. Thus, we can manipulate a balance factor $\gamma$ ($\beta=\gamma\beta'$) to make trade-off between the content and style matching.
\end{paragraph}

\begin{subsection}{Different Style Representations}

\begin{figure}[htpb]
\begin{center}
	\includegraphics[width=0.98\linewidth]{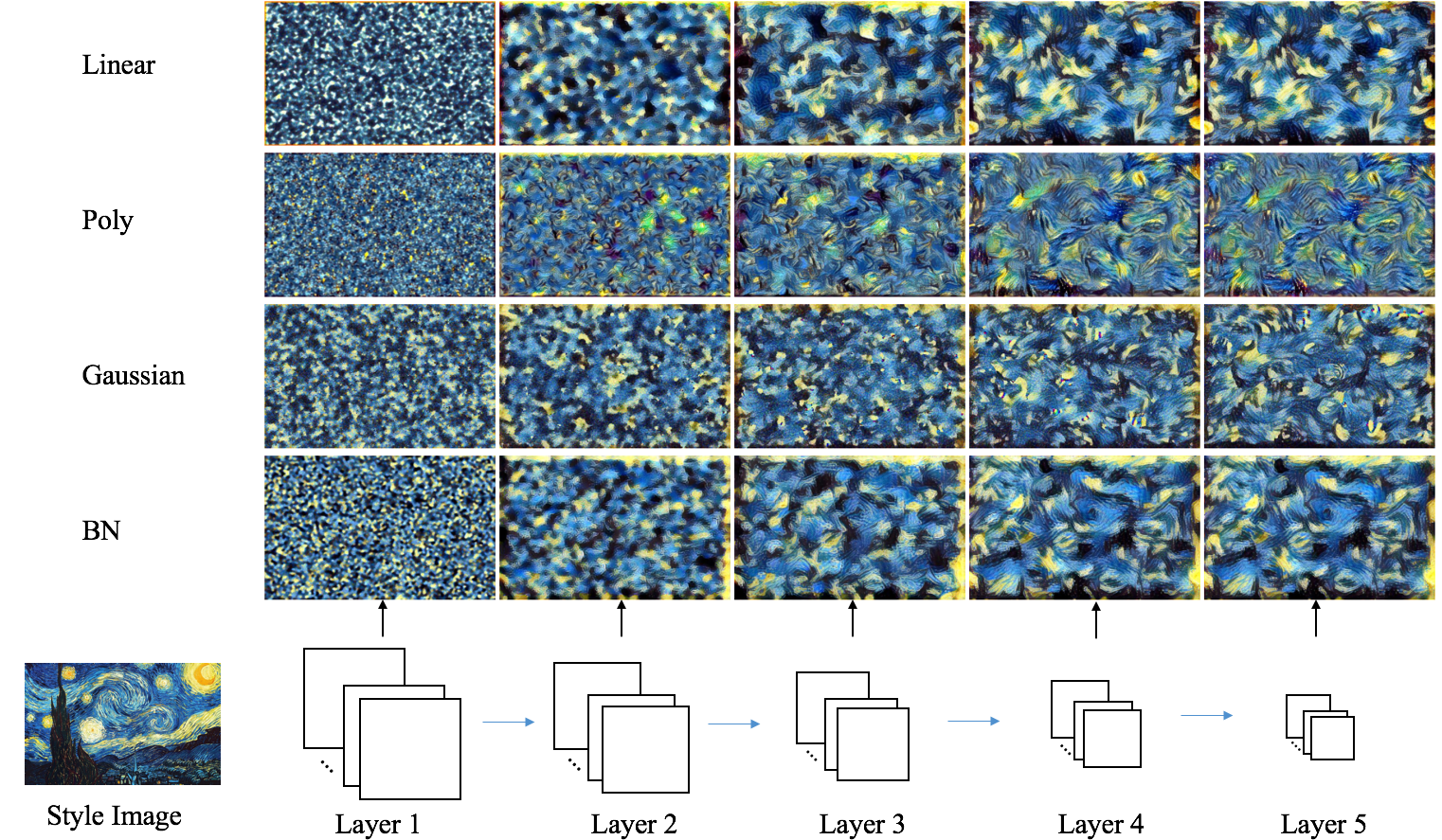}
\end{center}
\vspace{-2mm}
	\caption{Style reconstructions of different methods in five layers, respectively. Each row corresponds to one method and the reconstruction results are obtained by only using the style loss $\mathcal{L}_{style}$ with $\alpha=0$. We also reconstruct different style representations in different subsets of layers of VGG network. For example, layer 3 contains the style loss of the first 3 layers ($w_1=w_2=w_3=1.0$ and $w_4=w_5=0.0$).
	} \label{fig:style_reconstr}
\vspace{-2mm}
\end{figure}

\begin{figure*}[!htpb]
\begin{center}
	\subfigure{
	  \begin{overpic}[width=0.13\linewidth]{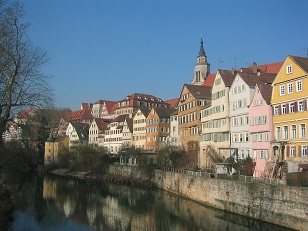}%
	    \put(-5,-5){\includegraphics[width=0.08\linewidth]{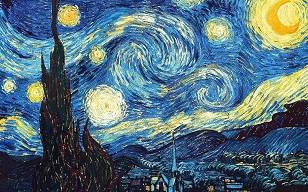}}
	  \end{overpic}
	}
	\subfigure{\includegraphics[width=0.13\linewidth]{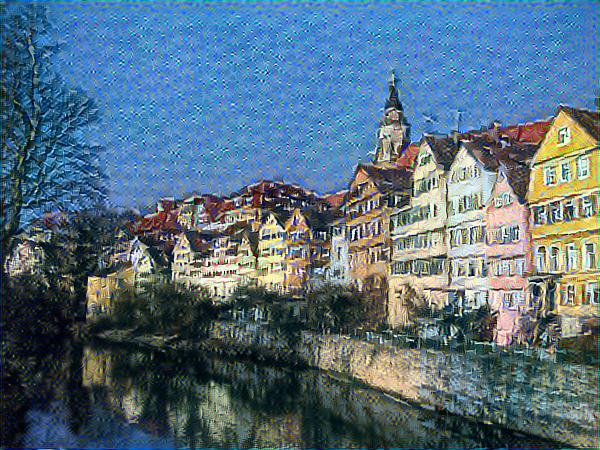}}~~
	\subfigure{\includegraphics[width=0.13\linewidth]{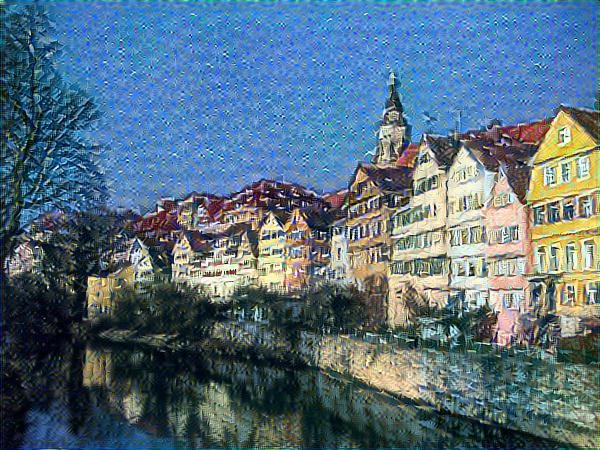}}~~
	\subfigure{\includegraphics[width=0.13\linewidth]{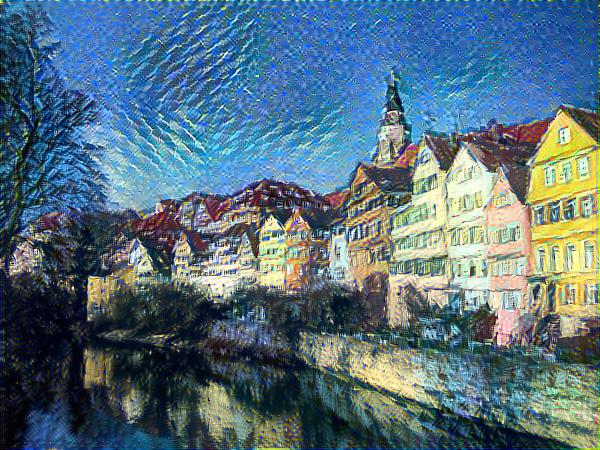}}~~
	\subfigure{\includegraphics[width=0.13\linewidth]{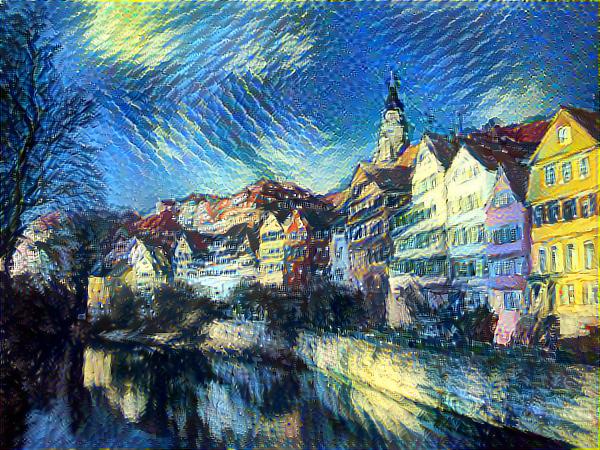}}~~
	\subfigure{\includegraphics[width=0.13\linewidth]{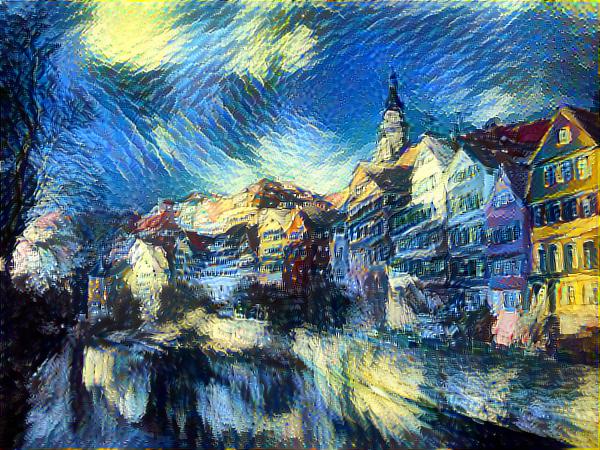}}\\
	\vspace{-1mm}
	\subfigure{
	  \begin{overpic}[width=0.13\linewidth]{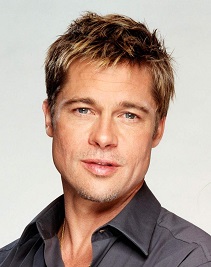}%
	    \put(-5,-5){\includegraphics[width=0.08\linewidth]{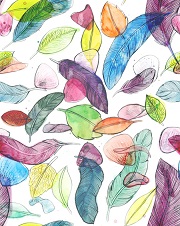}}
	  \end{overpic}
	}
	\subfigure{\includegraphics[width=0.13\linewidth]{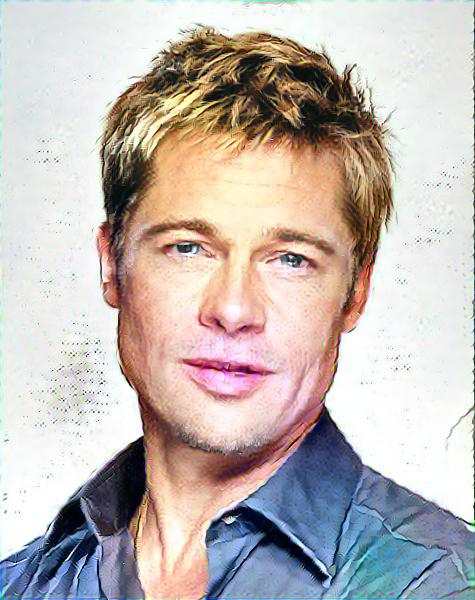}}~~
	\subfigure{\includegraphics[width=0.13\linewidth]{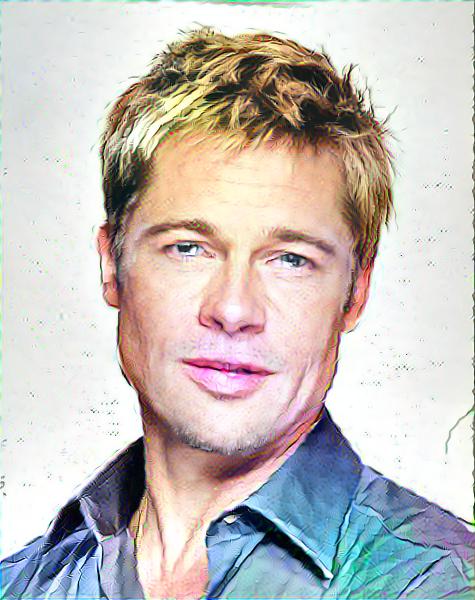}}~~
	\subfigure{\includegraphics[width=0.13\linewidth]{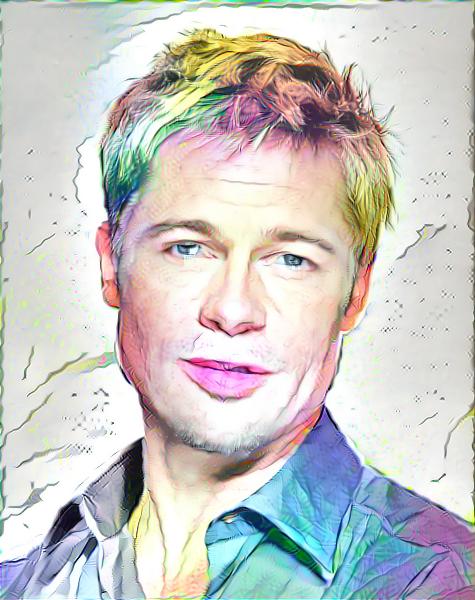}}~~
	\subfigure{\includegraphics[width=0.13\linewidth]{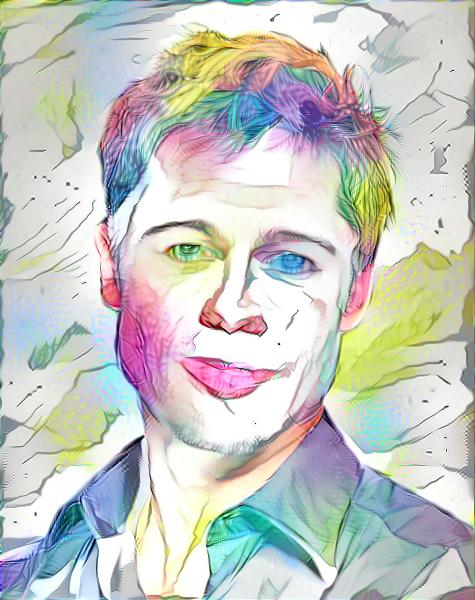}}~~
	\subfigure{\includegraphics[width=0.13\linewidth]{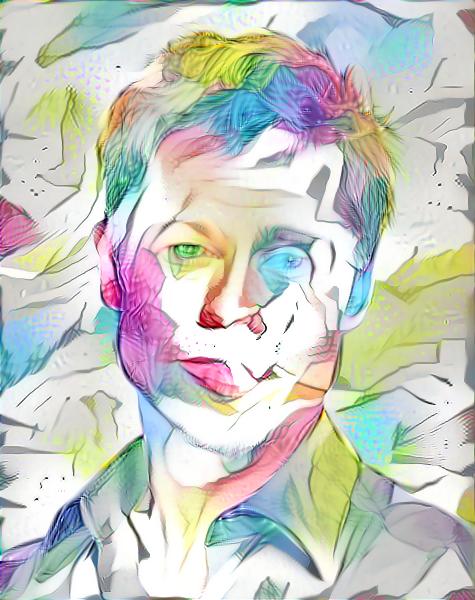}}\\
	\vspace{-1mm}
	\subfigure{
	  \begin{overpic}[width=0.13\linewidth]{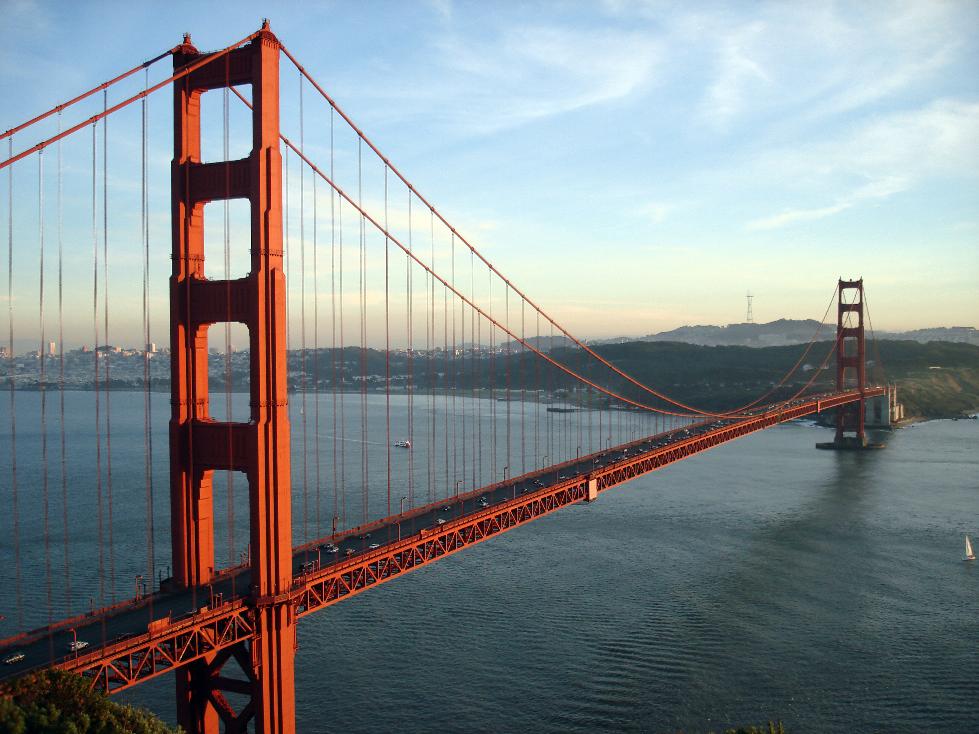}%
	    \put(-5,-5){\includegraphics[width=0.05\linewidth]{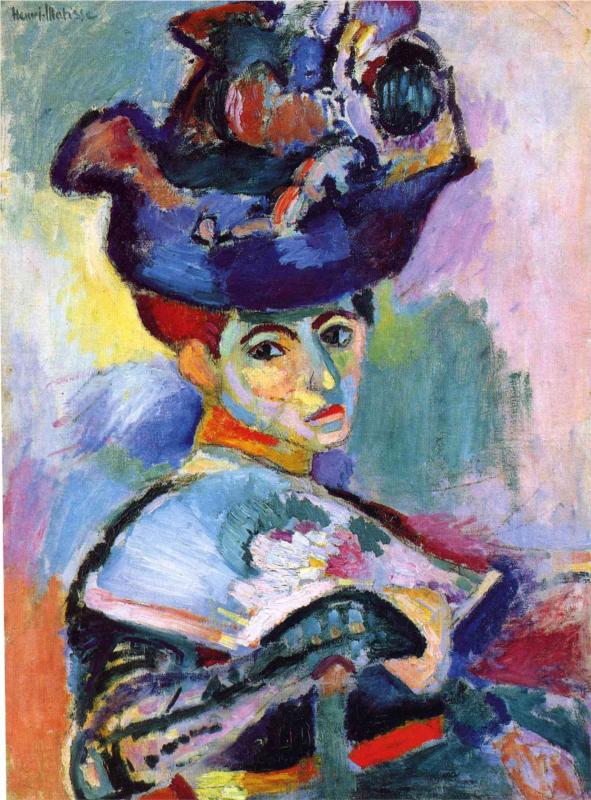}}
	  \end{overpic}
	}
	\subfigure{\includegraphics[width=0.13\linewidth]{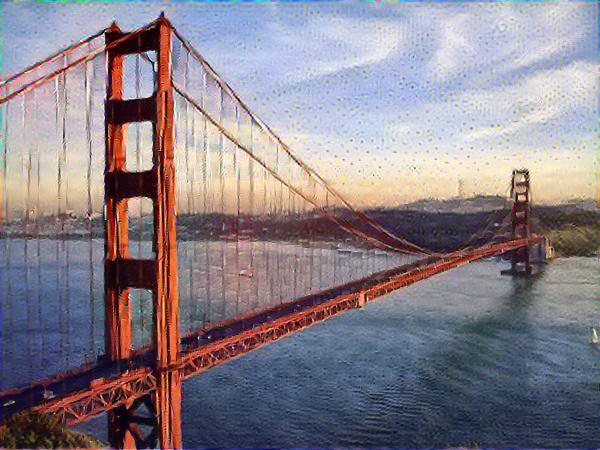}}~~
	\subfigure{\includegraphics[width=0.13\linewidth]{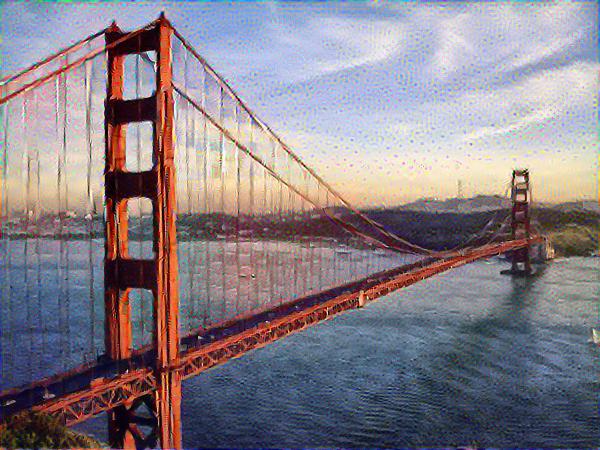}}~~
	\subfigure{\includegraphics[width=0.13\linewidth]{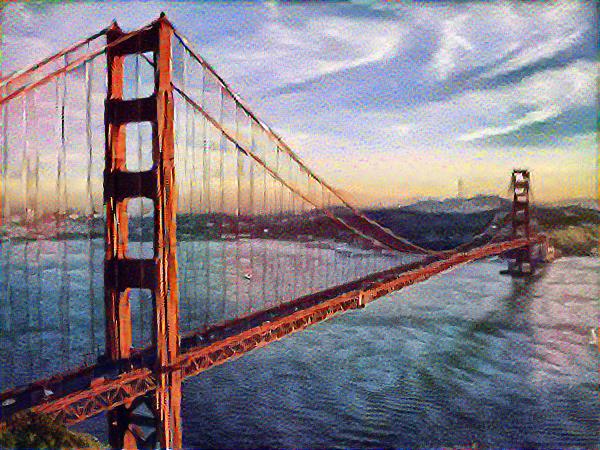}}~~
	\subfigure{\includegraphics[width=0.13\linewidth]{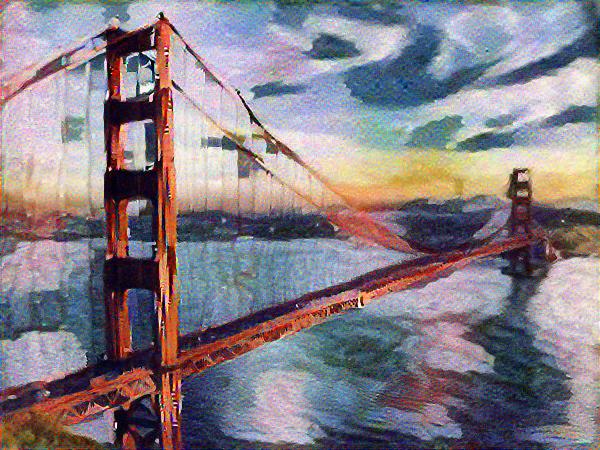}}~~
	\subfigure{\includegraphics[width=0.13\linewidth]{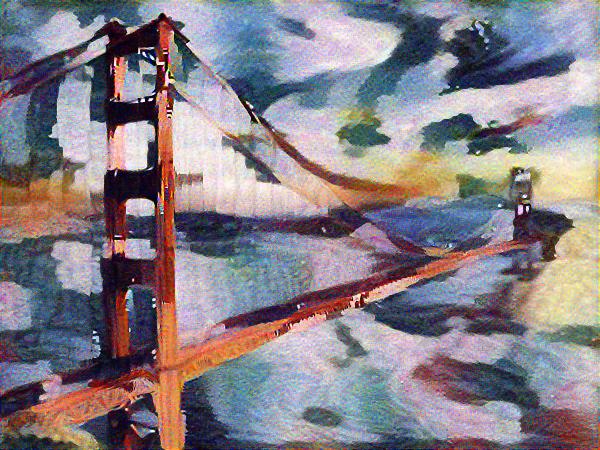}}\\
	\vspace{-1mm}
\setcounter{subfigure}{0}
	\subfigure[Content / Style]{
	  \begin{overpic}[width=0.13\linewidth]{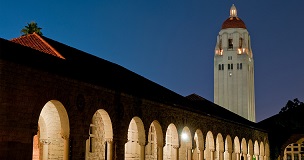}%
	    \put(-5,-5){\includegraphics[width=0.05\linewidth]{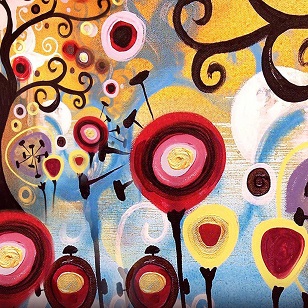}}
	  \end{overpic}
	}
	\subfigure[$\gamma=0.1$]{\includegraphics[width=0.13\linewidth]{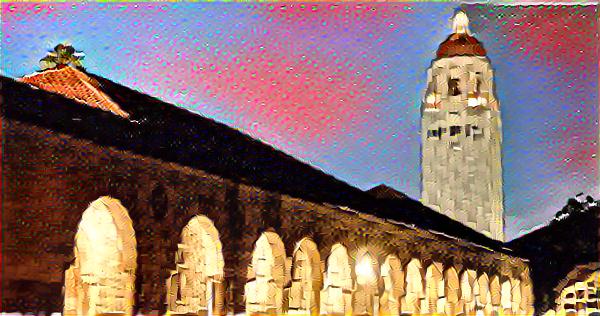}~~\label{fig:gamma0.1}}
	\subfigure[$\gamma=0.2$]{\includegraphics[width=0.13\linewidth]{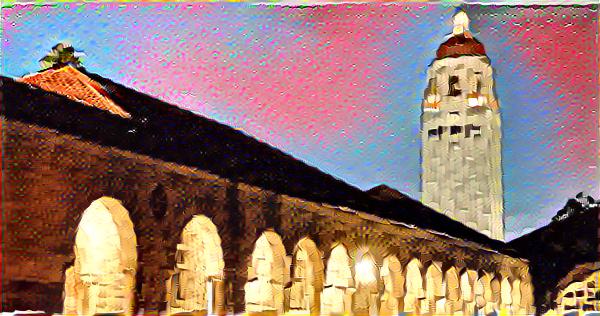}~~\label{fig:gamma0.2}}
	\subfigure[$\gamma=1.0$]{\includegraphics[width=0.13\linewidth]{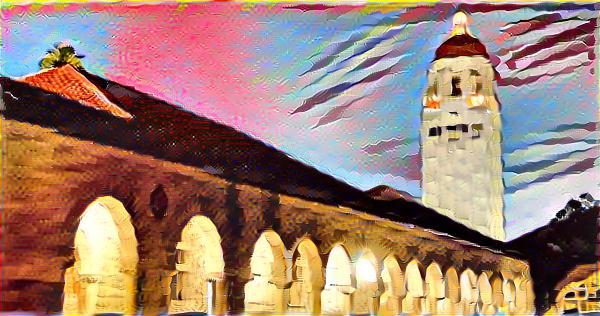}~~\label{fig:gamma1.0}}
	\subfigure[$\gamma=5.0$]{\includegraphics[width=0.13\linewidth]{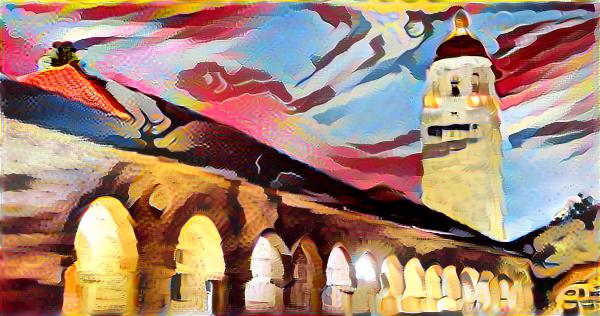}~~\label{fig:gamma5.0}}
	\subfigure[$\gamma=10.0$]{\includegraphics[width=0.13\linewidth]{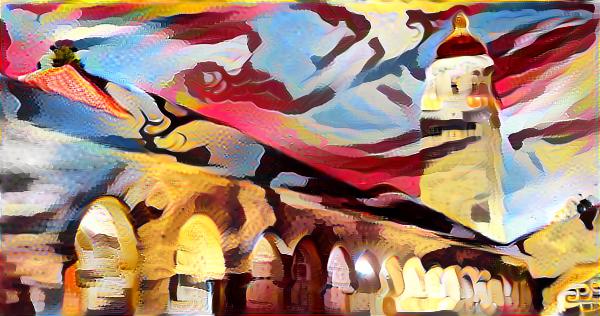}\label{fig:gamma10.0}}
\end{center}
	\caption{Results of the four methods (\emph{linear}, \emph{poly}, \emph{Gaussian} and \emph{BN}) with different balance factor $\gamma$. Larger $\gamma$ means more emphasis on the style loss.} \label{fig:effect_factor}
	\vspace{-2.5mm}
\end{figure*}

To validate that the extended neural style transfer methods can capture the style representation of an artistic image, we first visualize the style reconstruction results of different methods only using the style loss in Fig.~\ref{fig:style_reconstr}. Moreover, Fig.~\ref{fig:style_reconstr} also compares the style representations of different layers. On one hand, for a specific method (one row), the results show that different layers capture different levels of style: The textures in the top layers usually has larger granularity than those in the bottom layers. This is reasonable because each neuron in the top layers has larger receptive field and thus has the ability to capture more global textures. On the other hand, for a specific layer, Fig.~\ref{fig:style_reconstr} also demonstrates that the style captured by different methods differs. For example, in top layers, the textures captured by MMD with a linear kernel are composed by thick strokes. Contrarily, the textures captured by MMD with a polynomial kernel are more fine grained.

\end{subsection}

\begin{subsection}{Result Comparisons}
\begin{paragraph}{Effect of the Balance Factor}
We first explore the effect of the balance factor between the content loss and style loss by varying the weight $\gamma$. Fig.~\ref{fig:effect_factor} shows the results of four transfer methods with various $\gamma$ from $0.1$ to $10.0$. As intended, the global color information in the style image is successfully transfered to the content image, and the results with smaller $\gamma$ preserve more content details as shown in Fig.~\ref{fig:gamma0.1} and Fig.~\ref{fig:gamma0.2}. When $\gamma$ becomes larger, more stylized textures are incorporated into the results. For example, Fig.~\ref{fig:gamma5.0} and Fig.~\ref{fig:gamma10.0} have much more similar illumination and textures with the style image, while Fig.~\ref{fig:gamma1.0} shows a balanced result between the content and style. Thus, users can make trade-off between the content and the style by varying $\gamma$.

\end{paragraph}

\begin{figure}[htpb]
\begin{center}
\subfigure{
	  \begin{overpic}[width=0.185\linewidth]{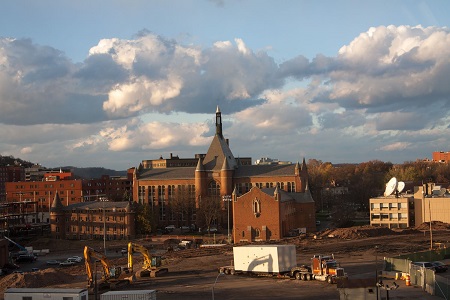}%
	    \put(-5,-5){\includegraphics[width=0.08\linewidth]{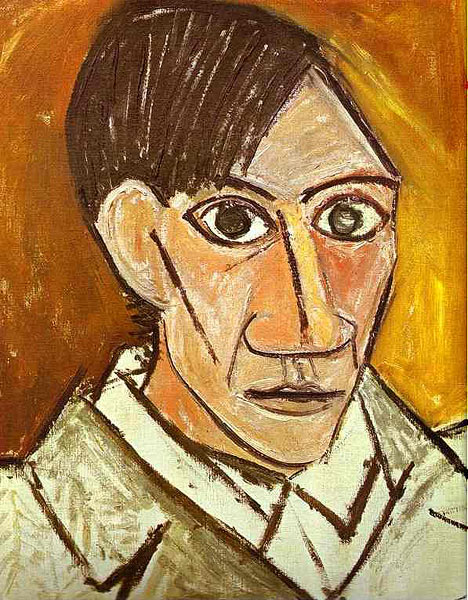}}
	  \end{overpic}
	}
	\subfigure{\includegraphics[width=0.185\linewidth]{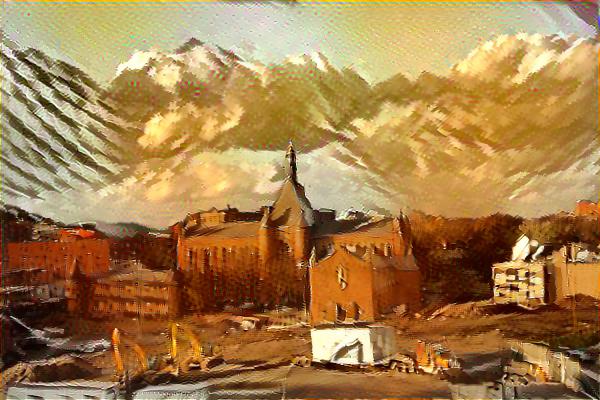}}
	\subfigure{\includegraphics[width=0.185\linewidth]{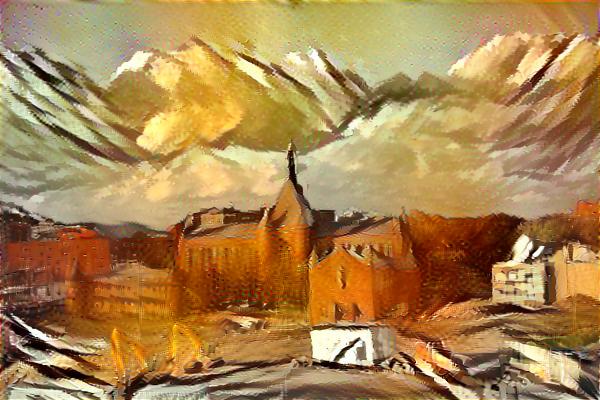}}
	\subfigure{\includegraphics[width=0.185\linewidth]{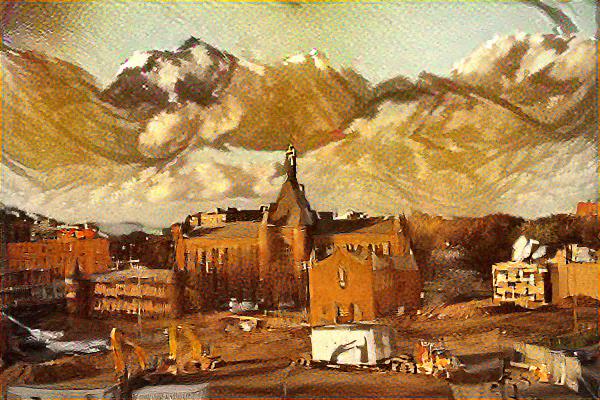}}
	\subfigure{\includegraphics[width=0.185\linewidth]{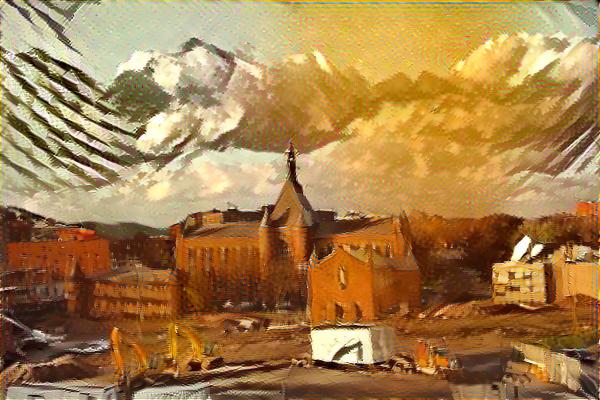}}\\
	\vspace{-1mm}

\subfigure{
	  \begin{overpic}[width=0.185\linewidth]{golden_gate.jpg}%
	    \put(-5,-5){\includegraphics[width=0.08\linewidth]{candy.jpg}}
	  \end{overpic}
	}
	\subfigure{\includegraphics[width=0.185\linewidth]{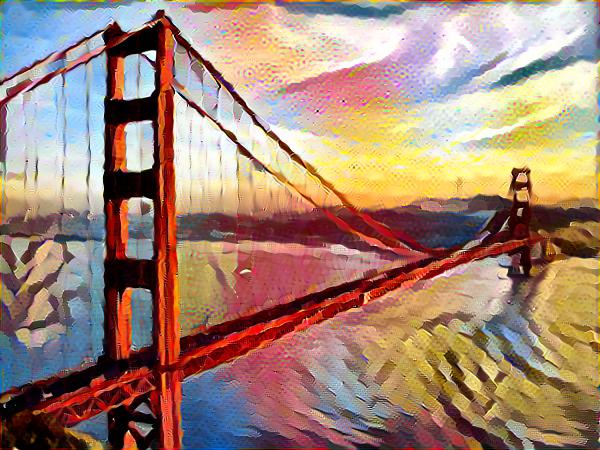}}
	\subfigure{\includegraphics[width=0.185\linewidth]{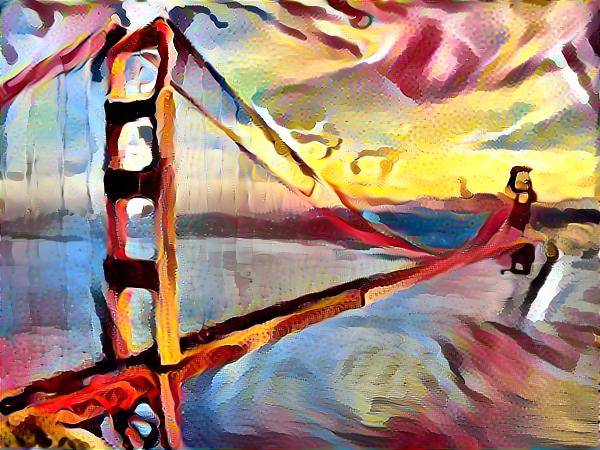}}
	\subfigure{\includegraphics[width=0.185\linewidth]{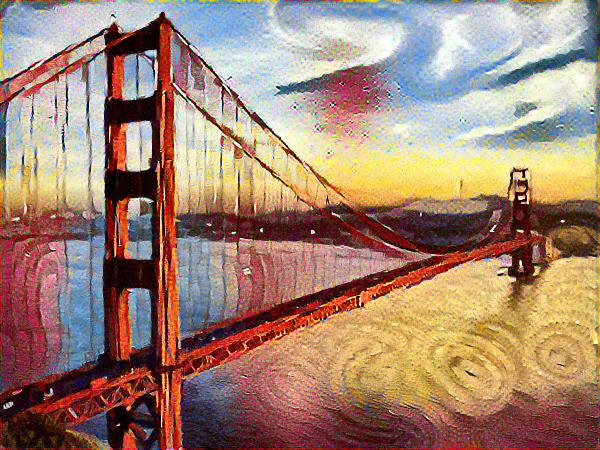}}
	\subfigure{\includegraphics[width=0.185\linewidth]{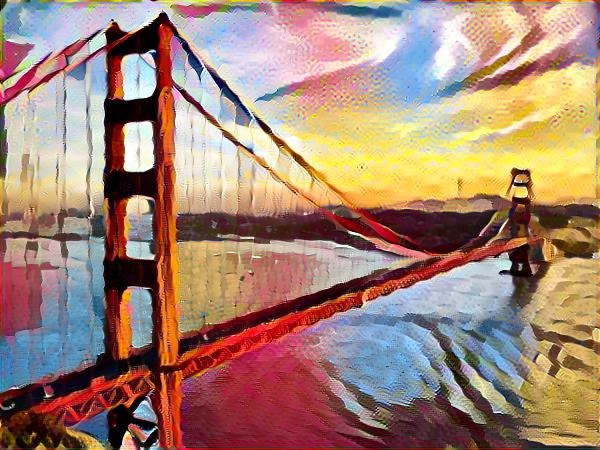}}\\
	\vspace{-1mm}

\subfigure{
	  \begin{overpic}[width=0.185\linewidth]{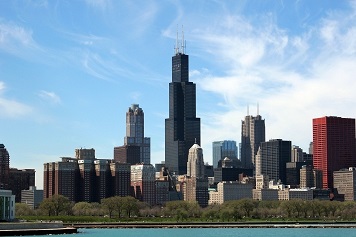}%
	    \put(-5,-5){\includegraphics[width=0.08\linewidth]{feathers.jpg}}
	  \end{overpic}
	}
	\subfigure{\includegraphics[width=0.185\linewidth]{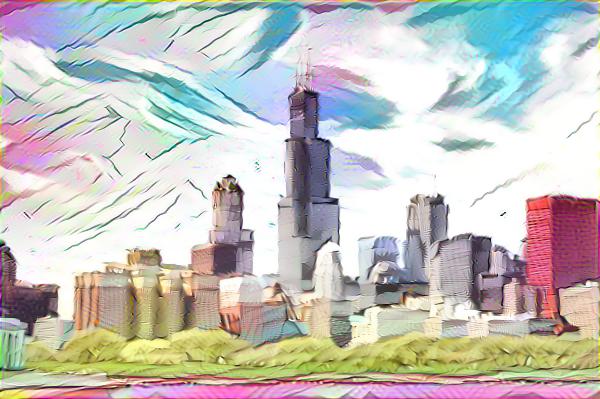}}
	\subfigure{\includegraphics[width=0.185\linewidth]{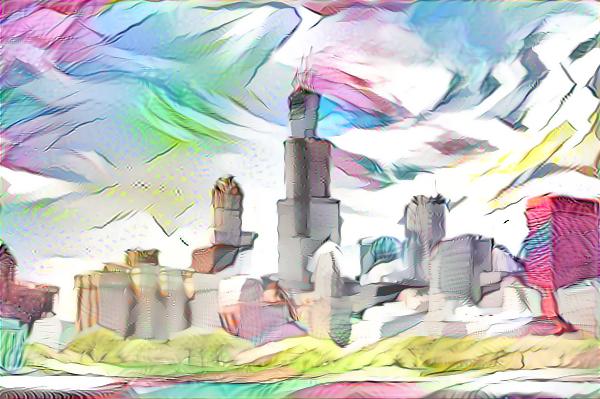}}
	\subfigure{\includegraphics[width=0.185\linewidth]{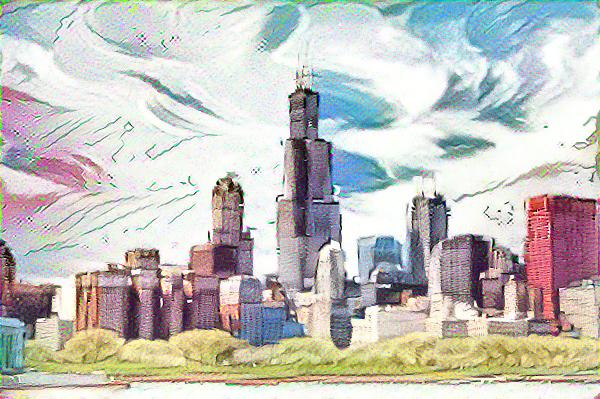}}
	\subfigure{\includegraphics[width=0.185\linewidth]{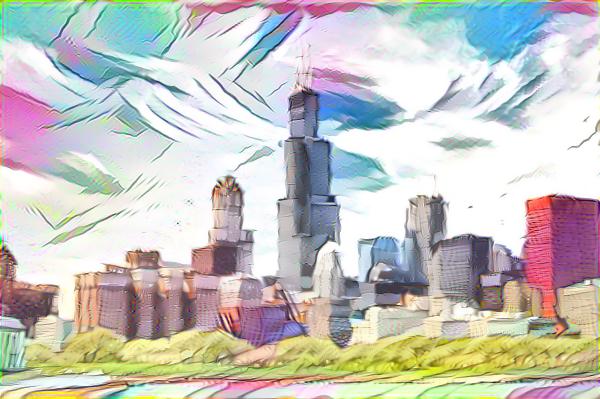}}\\
	\vspace{-1mm}

\subfigure{
	  \begin{overpic}[width=0.185\linewidth]{tubingen.jpg}%
	    \put(-5,-5){\includegraphics[width=0.08\linewidth]{woman-with-hat-matisse.jpg}}
	  \end{overpic}
	}
	\subfigure{\includegraphics[width=0.185\linewidth]{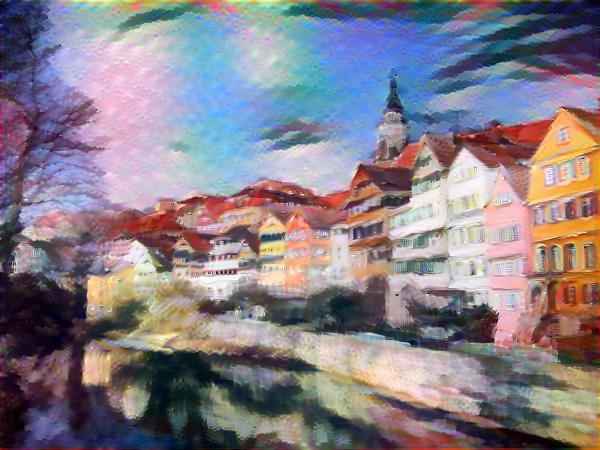}}
	\subfigure{\includegraphics[width=0.185\linewidth]{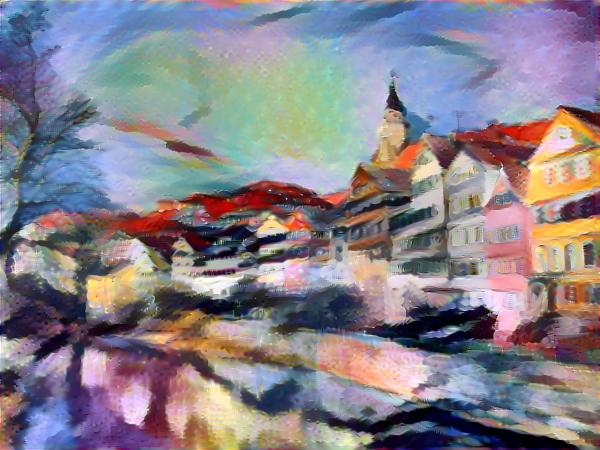}}
	\subfigure{\includegraphics[width=0.185\linewidth]{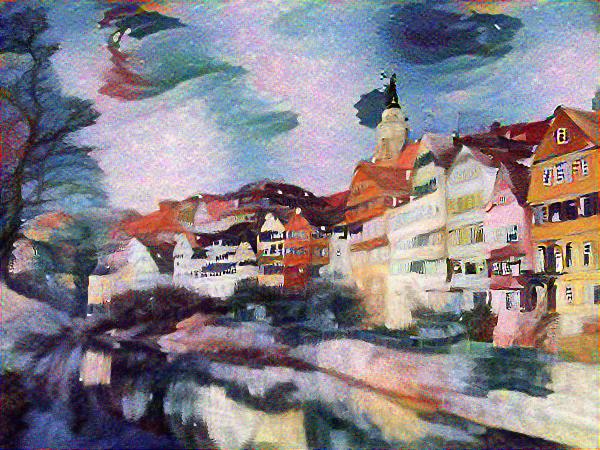}}
	\subfigure{\includegraphics[width=0.185\linewidth]{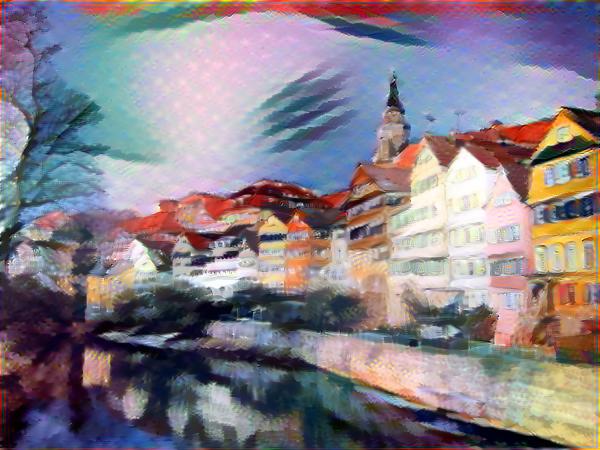}}\\
	\vspace{-1mm}

\subfigure{
	  \begin{overpic}[width=0.185\linewidth]{brad_pitt.jpg}%
	    \put(-5,-5){\includegraphics[width=0.08\linewidth]{starry_night.jpg}}
	  \end{overpic}
	}
	\subfigure{\includegraphics[width=0.185\linewidth]{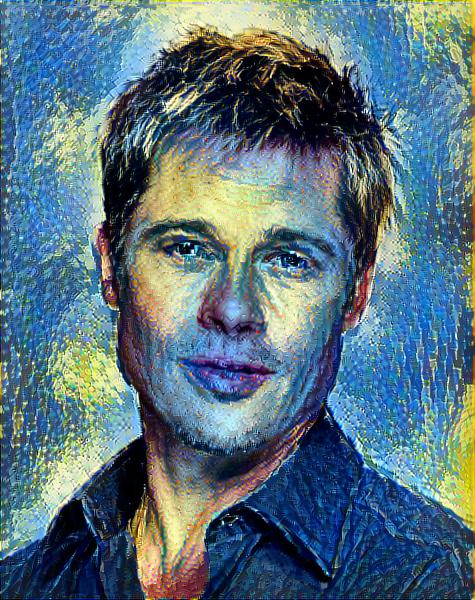}}
	\subfigure{\includegraphics[width=0.185\linewidth]{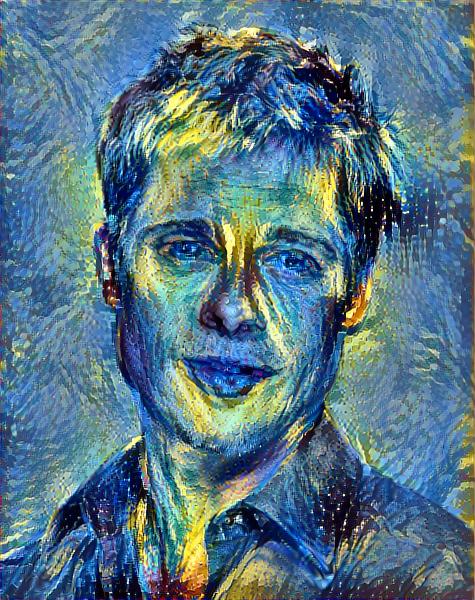}}
	\subfigure{\includegraphics[width=0.185\linewidth]{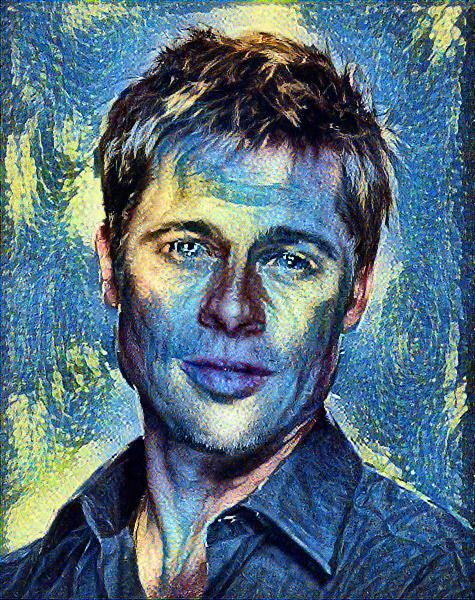}}
	\subfigure{\includegraphics[width=0.185\linewidth]{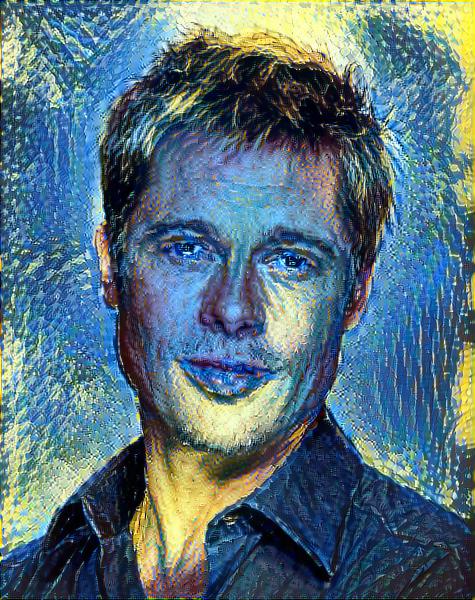}}\\
	\vspace{-1mm}

\setcounter{subfigure}{0}
\subfigure[Content / Style]{
	  \begin{overpic}[width=0.185\linewidth]{hoovertowernight.jpg}%
	    \put(-5,-5){\includegraphics[width=0.08\linewidth]{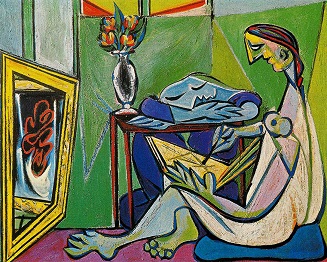}}
	  \end{overpic}
	}
	\subfigure[linear]{\includegraphics[width=0.185\linewidth]{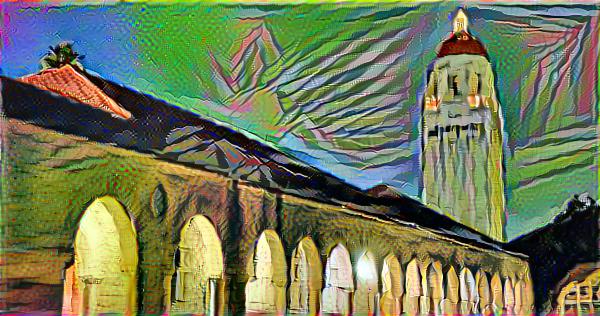}}
	\subfigure[poly]{\includegraphics[width=0.185\linewidth]{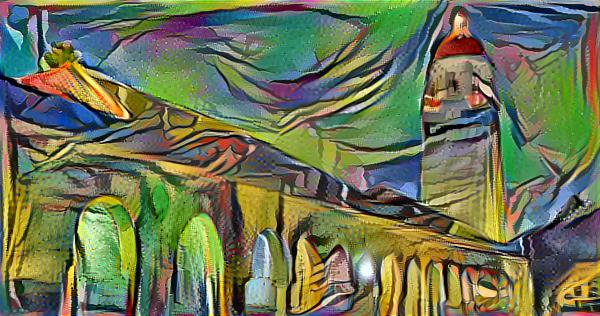}}
	\subfigure[Gaussian]{\includegraphics[width=0.185\linewidth]{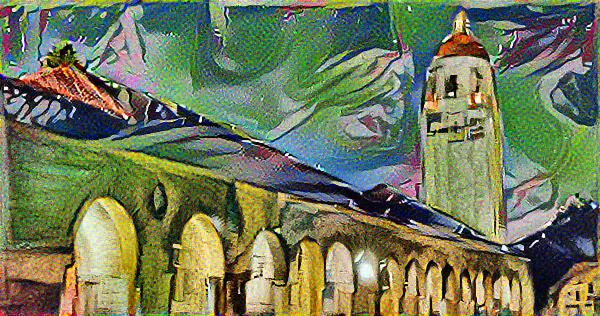}}
	\subfigure[BN]{\includegraphics[width=0.185\linewidth]{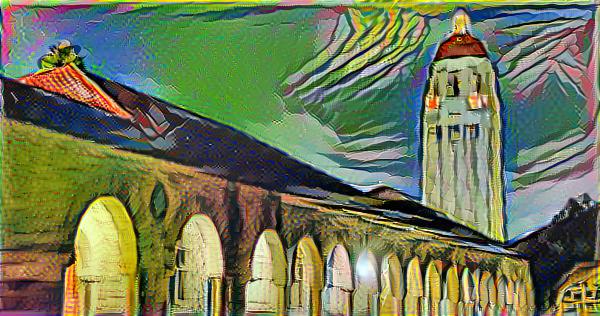}}\\
	\vspace{-1mm}

\end{center}
	\caption{Visual results of several style transfer methods, including \emph{linear}, \emph{poly}, \emph{Gaussian} and \emph{BN}. The balance factors $\gamma$ in the six examples are $2.0$, $2.0$, $2.0$, $5.0$, $5.0$ and $5.0$, respectively.} \label{fig:vis_results}
\end{figure}

\begin{figure*}[htpb]
\begin{center}
	\subfigure{
	  \begin{overpic}[width=0.14\linewidth]{hoovertowernight.jpg}%
	    \put(-5,-5){\includegraphics[width=0.08\linewidth]{starry_night.jpg}}
	  \end{overpic}
	}
	\subfigure{\includegraphics[width=0.14\linewidth]{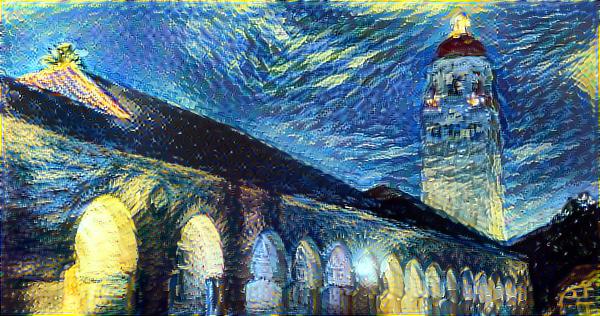}}
	\subfigure{\includegraphics[width=0.14\linewidth]{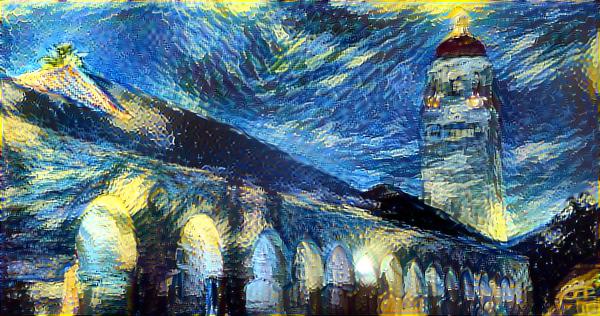}}
	\subfigure{\includegraphics[width=0.14\linewidth]{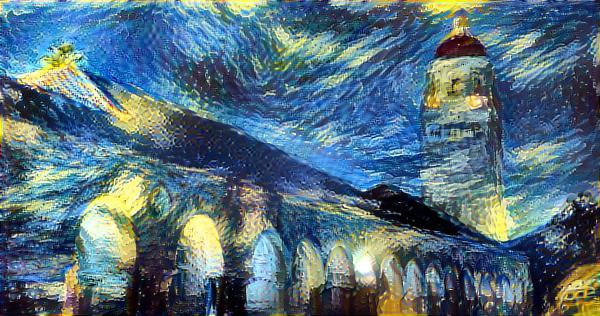}}
	\subfigure{\includegraphics[width=0.14\linewidth]{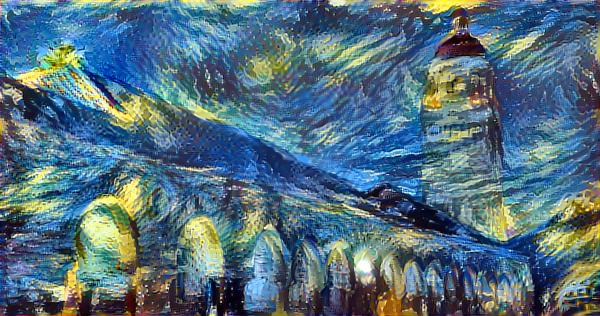}}
	\subfigure{\includegraphics[width=0.14\linewidth]{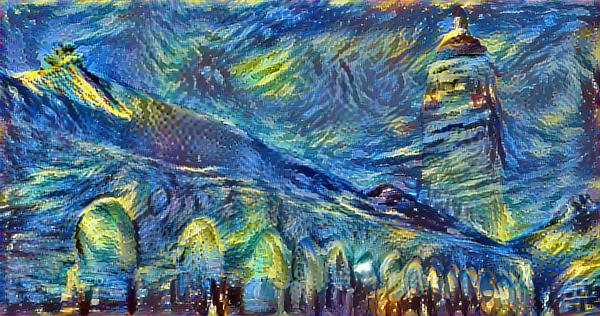}}\\
	\vspace{-1mm}
	\subfigure{
	  \begin{overpic}[width=0.14\linewidth]{brad_pitt.jpg}%
	    \put(-5,-5){\includegraphics[width=0.08\linewidth]{candy.jpg}}
	  \end{overpic}
	}
	\subfigure{\includegraphics[width=0.14\linewidth]{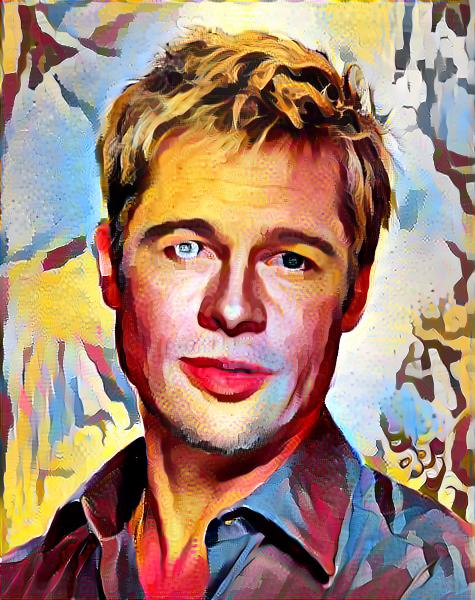}}
	\subfigure{\includegraphics[width=0.14\linewidth]{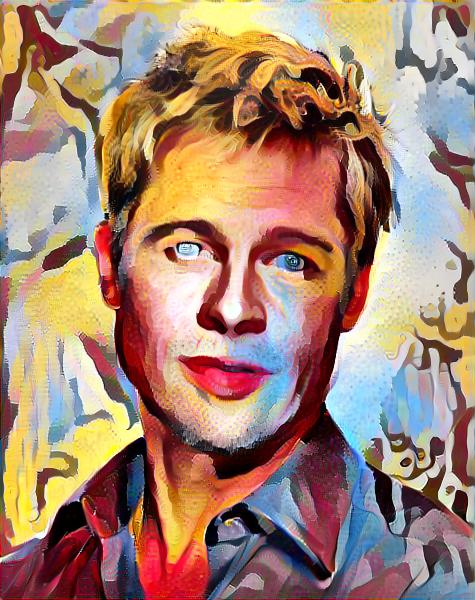}}
	\subfigure{\includegraphics[width=0.14\linewidth]{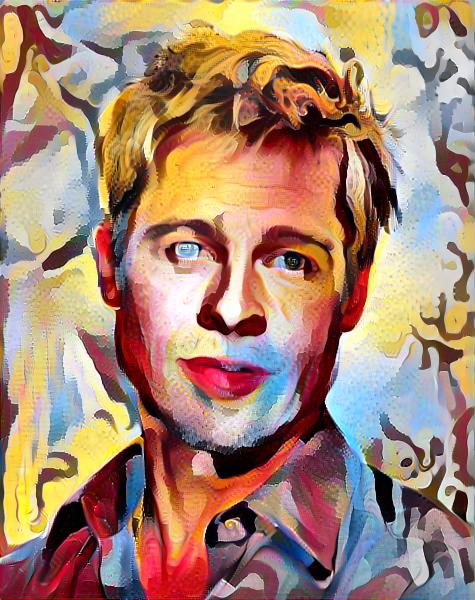}}
	\subfigure{\includegraphics[width=0.14\linewidth]{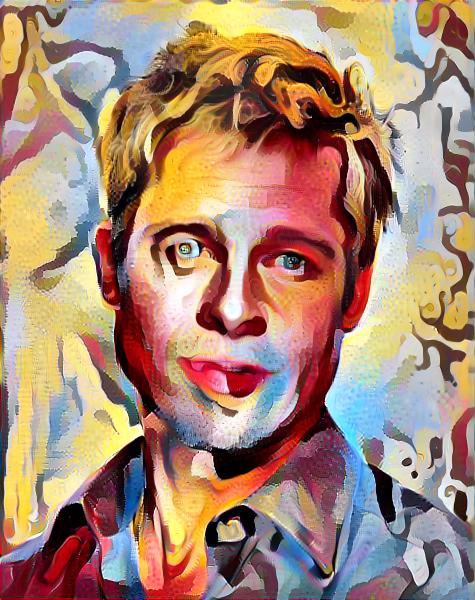}}
	\subfigure{\includegraphics[width=0.14\linewidth]{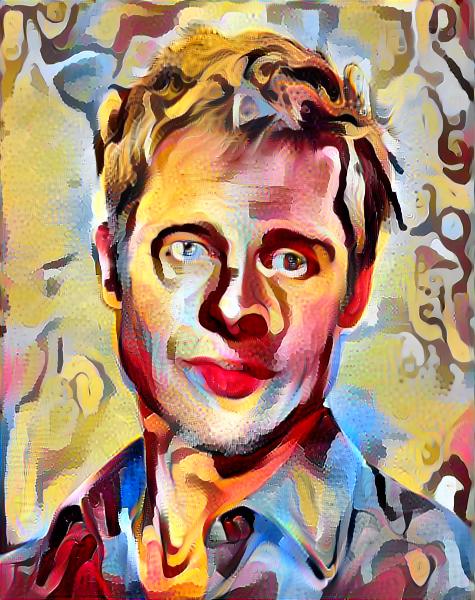}}\\
	\vspace{-1mm}
	\subfigure{
	  \begin{overpic}[width=0.14\linewidth]{chicago.jpg}%
	    \put(-5,-5){\includegraphics[width=0.08\linewidth]{feathers.jpg}}
	  \end{overpic}
	}
	\subfigure{\includegraphics[width=0.14\linewidth]{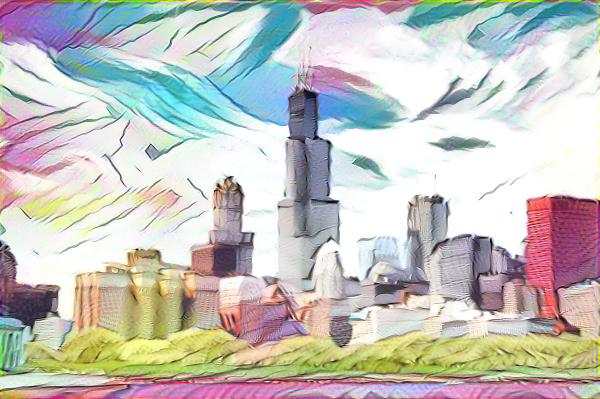}}
	\subfigure{\includegraphics[width=0.14\linewidth]{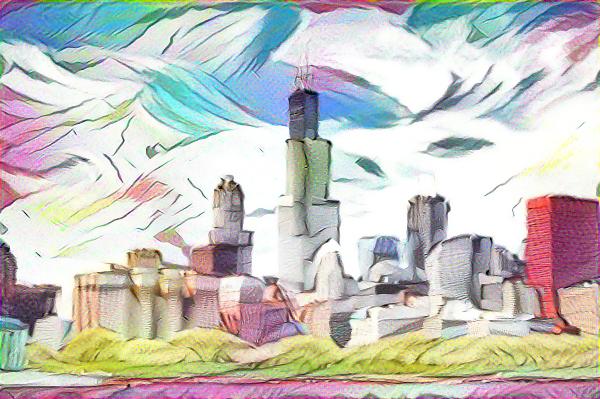}}
	\subfigure{\includegraphics[width=0.14\linewidth]{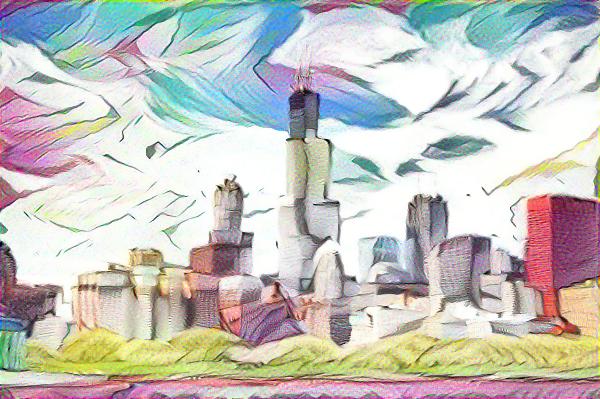}}
	\subfigure{\includegraphics[width=0.14\linewidth]{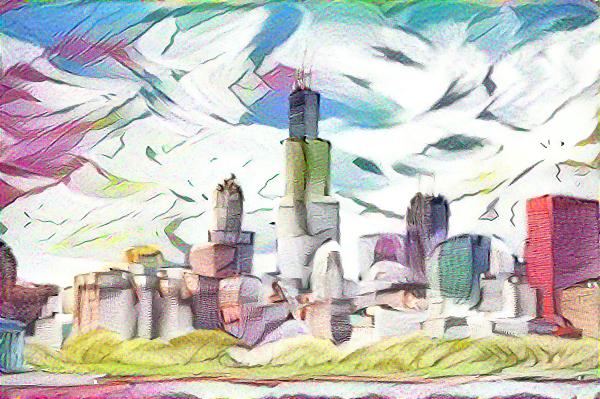}}
	\subfigure{\includegraphics[width=0.14\linewidth]{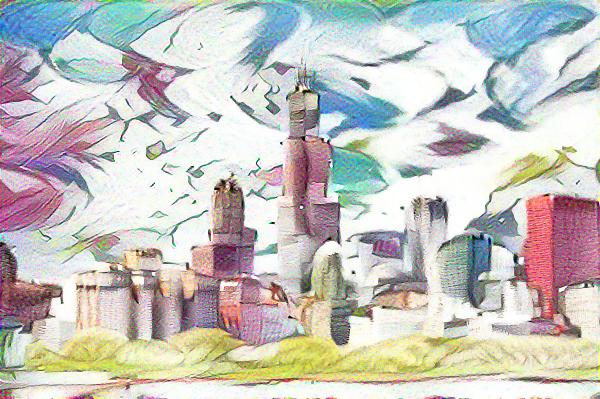}}\\
\setcounter{subfigure}{0}
	\subfigure[Content / Style]{
	  \begin{overpic}[width=0.14\linewidth]{tubingen.jpg}%
	    \put(-5,-5){\includegraphics[width=0.05\linewidth]{woman-with-hat-matisse.jpg}}
	  \end{overpic}
	}
	\subfigure[(0.9, 0.1)]{\includegraphics[width=0.14\linewidth]{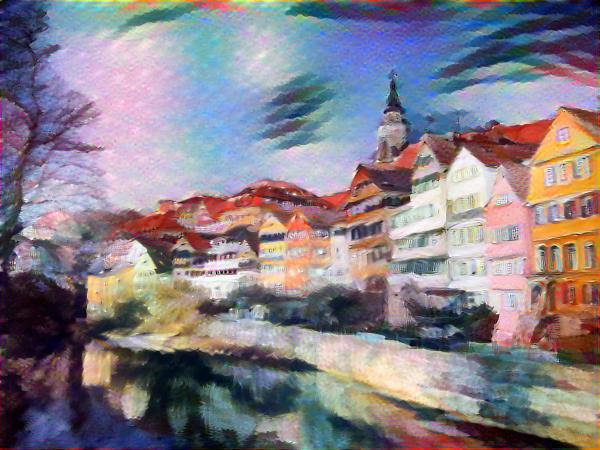}\label{fig:multi0.9}}
	\subfigure[(0.7, 0.3)]{\includegraphics[width=0.14\linewidth]{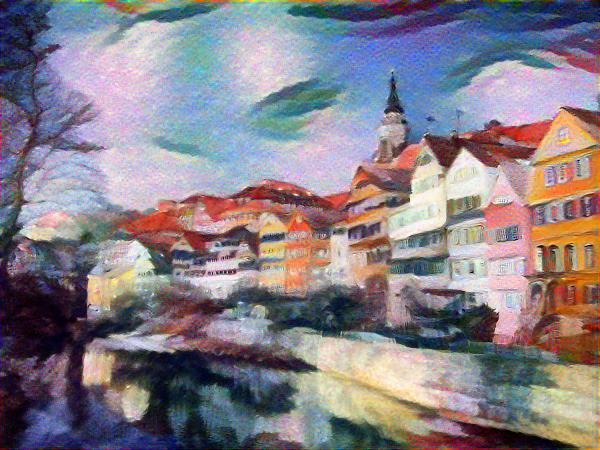}\label{fig:multi0.7}}
	\subfigure[(0.5, 0.5)]{\includegraphics[width=0.14\linewidth]{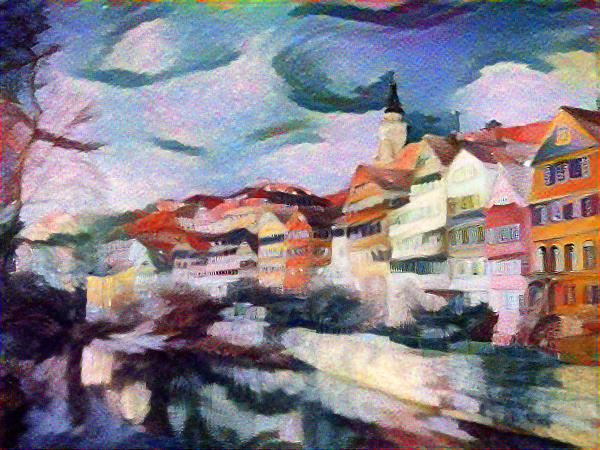}\label{fig:multi0.5}}
	\subfigure[(0.3, 0.7)]{\includegraphics[width=0.14\linewidth]{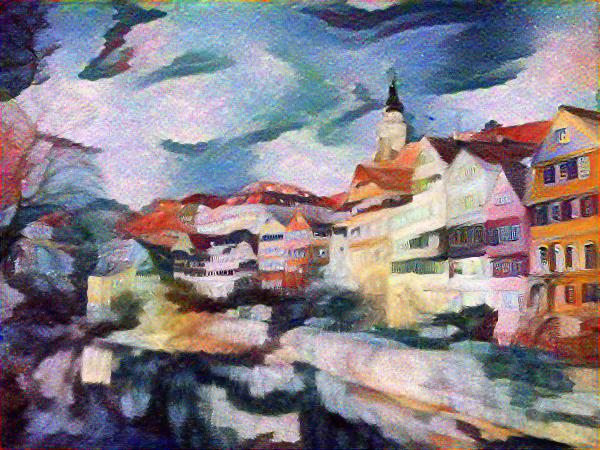}\label{fig:multi0.3}}
	\subfigure[(0.1, 0.9)]{\includegraphics[width=0.14\linewidth]{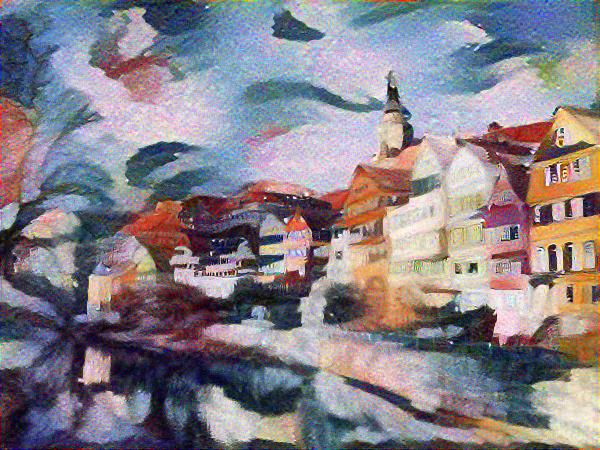}\label{fig:multi0.1}}
\end{center}
\vspace{-1.5mm}
	\caption{Results of two fusion methods: \emph{BN} + \emph{poly} and \emph{linear} + \emph{Gaussian}. The top two rows are the results of first fusion method and the bottom two rows correspond to the second one. Each column shows the results of a  balance weight between the two methods. $\gamma$ is set as 5.0.} \label{fig:multi}
	\vspace{-2.5mm}
\end{figure*}

\begin{paragraph}{Comparisons of Different Transfer Methods}
Fig.~\ref{fig:vis_results} presents the results of various pairs of content and style images with different transfer methods\footnote{More results can be found at\\ {{http://www.icst.pku.edu.cn/struct/Projects/mmdstyle/result-1000/show-full.html}}}. Similar to matching Gram matrices, which is equivalent to the \emph{poly} method, the other three methods can also transfer satisfied styles from the specified style images. This empirically demonstrates the correctness of our interpretation of neural style transfer: Style transfer is essentially a domain adaptation problem, which aligns the feature distributions. Particularly, when the weight on the style loss becomes higher (namely, larger $\gamma$), the differences among the four methods are getting larger. This indicates that these methods implicitly capture different aspects of style, which has also been shown in Fig.~\ref{fig:style_reconstr}. Since these methods have their unique properties, they could provide more choices for users to stylize the content image. For example, \emph{linear} achieves comparable results with other methods, yet requires lower computation complexity.

\end{paragraph}

\begin{paragraph}{Fusion of Different Neural Style Transfer Methods}
Since we have several different neural style transfer methods, we propose to combine them to produce new transfer results. Fig.~\ref{fig:multi} demonstrates the fusion results of two combinations (\emph{linear} + \emph{Gaussian} and \emph{poly} + \emph{BN}). Each row presents the results with different balance between the two methods. For example, Fig.~\ref{fig:multi0.9} in the first two rows emphasize more on \emph{BN} and Fig.~\ref{fig:multi0.1} emphasizes more on \emph{poly}. The results in the middle columns show the interpolation between these two methods. We can see that the styles of different methods are blended well using our method.

\end{paragraph}

\end{subsection}

\end{section}

%% file: src/conclusion.tex
\begin{section}{Conclusion}
Despite the great success of neural style transfer, the rationale behind neural style transfer was far from crystal. The vital ``trick'' for style transfer is to match the Gram matrices of the features in a layer of a CNN. Nevertheless, subsequent literatures about neural style transfer just directly improves upon it without investigating it in depth. In this paper, we present a timely explanation and interpretation for it. First, we theoretically prove that matching the Gram matrices is equivalent to a specific Maximum Mean Discrepancy (MMD) process. Thus, the style information in neural style transfer is intrinsically represented by the  distributions of activations in a CNN, and the style transfer can be achieved by distribution alignment. Moreover, we exploit several other distribution alignment methods, and find that these methods all yield promising transfer results. Thus, we justify the claim that neural style transfer is essentially a special domain adaptation problem both theoretically and empirically. We believe this interpretation provide a new lens to re-examine the style transfer problem, and will inspire more exciting works in this research area.

\end{section}